\documentclass[letterpaper, 10 pt, conference]{ieeeconf}  

\IEEEoverridecommandlockouts                              

\usepackage{multirow}
\usepackage{cite}
\usepackage{amsmath,amssymb,amsfonts}
\usepackage{algorithmic}
\usepackage{graphicx}
\usepackage{cuted}
\usepackage{capt-of}
\usepackage{float}
\usepackage{booktabs}
\usepackage{threeparttable}
\usepackage{mathtools}
\usepackage[mathscr]{eucal}
\usepackage[table,xcdraw]{xcolor}
\usepackage{colortbl}
\usepackage{balance}
\usepackage{textcomp}
\usepackage{marvosym}
\usepackage{algorithmic}
\usepackage{indentfirst}
\definecolor{tabfirst}{rgb}{1, 0.7, 0.7} 
\definecolor{tabsecond}{rgb}{1, 0.85, 0.7} 
\definecolor{tabthird}{rgb}{1, 1, 0.7} 

\makeatletter
\let\NAT@parse\undefined
\makeatother
\usepackage{hyperref}

\title{CRUISE: \underline{C}ooperative \underline{R}econstr\underline{u}ction and Editing \underline{i}n V2X \underline{S}c\underline{e}narios using Gaussian Splatting
}

\author{Haoran Xu$^{1,2,7*}$, Saining Zhang$^{1,3*}$, Peishuo Li$^{3*}$, Baijun Ye$^{4}$, Xiaoxue Chen$^{1}$, \\Huan-ang Gao$^{1}$, Jv Zheng$^{1}$, Xiaowei Song$^{1}$, Ziqiao Peng$^{5}$, Run Miao$^{6}$, Jinrang Jia$^{7}$, \\Yifeng Shi$^{7}$, Guangqi Yi$^{7}$, Hang Zhao$^{4}$, Hao Tang$^{8}$, Hongyang Li$^{9}$, Kaicheng Yu$^{10}$, Hao Zhao$^{1,11\dag}$
\thanks{* Equal contribution. $^{1}$Institute for AI Industry Research (AIR), Tsinghua University; $^{2}$Beijing Institute of Technology; $^{3}$Nanyang Technological University; $^{4}$Tsinghua University; $^{5}$Renmin University of China; $^{6}$Beijing University of Technology; $^{7}$Baidu Inc; $^{8}$Peking University; $^{9}$Shanghai AI Lab;$^{10}$Westlake University; $^{11}$Beijing Academy of Artificial Intelligence.}
\thanks{$\dag$ Corresponding author. zhaohao@air.tsinghua.edu.cn}%
}

\begin{document}

\maketitle
\thispagestyle{empty}
\pagestyle{empty}

\begin{abstract}
Vehicle-to-everything (V2X) communication plays a crucial role in autonomous driving, enabling cooperation between vehicles and infrastructure. While simulation has significantly contributed to various autonomous driving tasks, its potential for data generation and augmentation in V2X scenarios remains underexplored. In this paper, we introduce CRUISE, a comprehensive reconstruction-and-synthesis framework designed for V2X driving environments. CRUISE employs decomposed Gaussian Splatting to accurately reconstruct real-world scenes while supporting flexible editing. By decomposing dynamic traffic participants into editable Gaussian representations, CRUISE allows for seamless modification and augmentation of driving scenes. Furthermore, the framework renders images from both ego-vehicle and infrastructure views, enabling large-scale V2X dataset augmentation for training and evaluation. Our experimental results demonstrate that: 1) CRUISE reconstructs real-world V2X driving scenes with high fidelity; 2) using CRUISE improves 3D detection across ego-vehicle, infrastructure, and cooperative views, as well as cooperative 3D tracking on the V2X-Seq benchmark; and 3) CRUISE effectively generates challenging corner cases. The code will be publicly available at \url{https://github.com/SainingZhang/CRUISE}.

\end{abstract}
\section{Introduction}
Autonomous driving has been advancing rapidly, offering the potential to revolutionize transportation by enhancing traffic safety and efficiency. As end-to-end autonomous driving models continue to emerge, the need for scalable, domain-gap-free simulations capable of supporting closed-loop, real-world evaluations has become increasingly apparent. Meanwhile, effective simulation frameworks are considered essential for training robust autonomous systems and ensuring their reliability across diverse driving conditions~\cite{wang2025unifying,xiao2025simulate,xu2025challenger,guo2025dist,li2025avd2,li2025uniscene,gao2024scp}.

In recent years, Neural Radiance Fields (NeRFs) \cite{mildenhall2020nerf,barron2021mipnerf,yuan2024slimmerf,liu2024rip} and Gaussian Splatting (GS) \cite{kerbl3Dgaussians} have emerged as fundamental techniques for high-fidelity 3D scene reconstruction. In the context of street-scene simulation, these methods~\cite{tancik2022block,martin2021nerf,rematas2022urban,guo2023streetsurf,yan2024oasim,zhang2024drone} have primarily focused on reconstructing static scenes, often overlooking dynamic elements such as moving vehicles. More recent approaches \cite{yang2023emernerf,turki2023suds,ost2021neural,kundu2022panoptic,yang2023unisim,wu2023mars,tonderski2024neurad,ml-nsg,zhou2023drivinggaussian,yan2024street,huang2024s3gaussian} have introduced 4D reconstruction techniques that effectively capture both dynamic traffic participants and static backgrounds with high fidelity.

\begin{figure}
  \centering
  \includegraphics[width=0.485\textwidth]{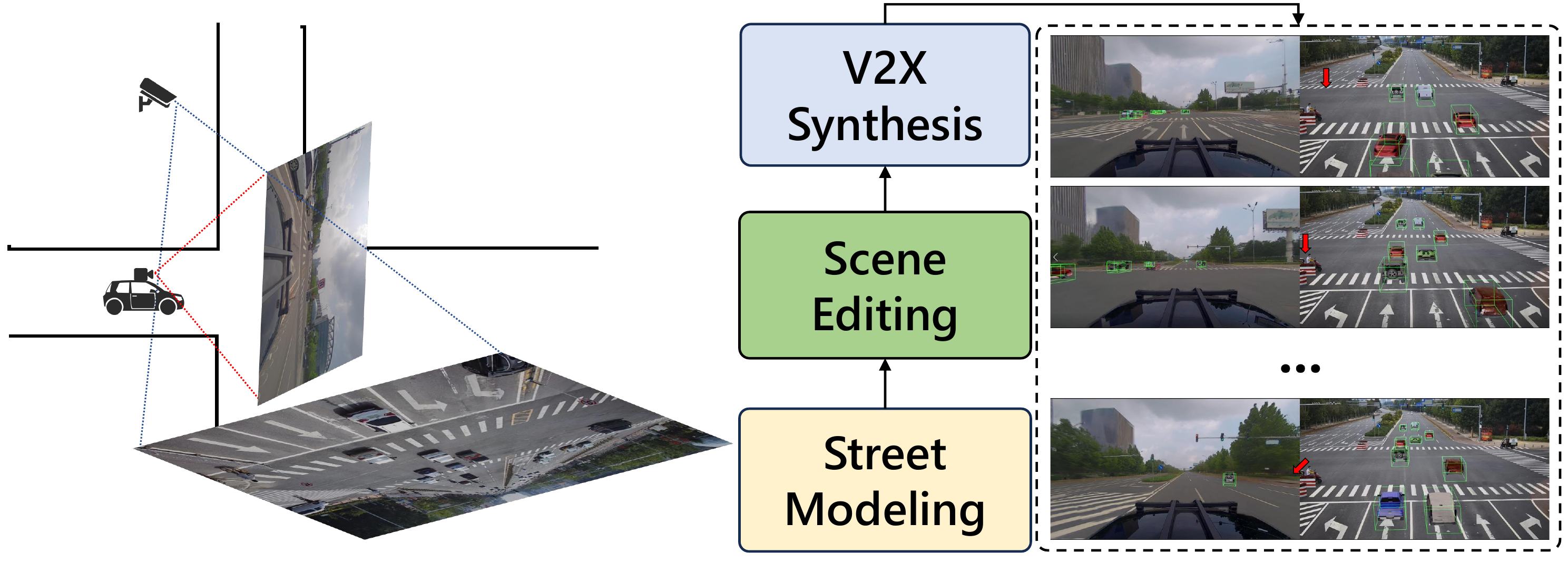}
  \vspace{-7mm}
  \caption{On the left is the V2X setup, showing captures from both the ego vehicle and the infrastructure. On the right is the CRUISE pipeline, which progresses from street modeling to scene editing and, ultimately, to V2X data synthesis. The red arrow indicates the ego‑vehicle’s position.}
  \label{fig:teaser}
  \vspace{-6mm}
\end{figure}

While the aforementioned studies have achieved remarkable simulation results using ego‑vehicle views, the future of end‑to‑end autonomous driving will likely require perspectives beyond a car‑centric approach. Recently, V2X communication, which integrates sensor data from both ego-vehicles and infrastructure, has emerged as a promising paradigm for enhancing autonomous driving capabilities \cite{d2024autonomous,yuan2023autonomous}. Numerous works have explored V2X-enabled tasks, including detection \cite{WangV2vnet:ECCV20,HuCollaboration:CVPR23,yu2023flow,WangUMC:ICCV23,qiu2022distributed}, tracking \cite{YuV2XSeq:CVPR23}, segmentation \cite{xu2022cobevt}, localization \cite{jiang2023roadside,dong2023lidar}, and forecasting \cite{YuV2XSeq:CVPR23,ruan2023learning,song2024collaborative}. However, these efforts predominantly focus on task-specific methodologies while overlooking the need for a closed-loop data ecosystem to support V2X research and deployment. In this work, we introduce CRUISE, the first GS-based V2X simulation framework. CRUISE reconstructs real-world driving environments with high fidelity and diversifies the V2X datasets.

To enable efficient scene editing and data generation, CRUISE first reconstructs the environment from V2X images using decomposed GS, which effectively separates dynamic vehicles from static street scenes. Building upon this reconstruction, we introduce a generative editing paradigm that allows for the seamless modification of traffic scenarios by synthesizing vehicle Gaussian assets and adjusting scene composition. Once edited, the reconstructed scenes can be rendered from both ego-vehicle and infrastructure perspectives, producing high-fidelity V2X datasets for downstream tasks, as illustrated in Fig. \ref{fig:teaser}.

To evaluate the effectiveness of our data generation framework, we validate CRUISE using the V2X-Seq dataset \cite{yu2023v2x} as a data augmentation strategy for multiple 3D detection and cooperative 3D tracking tasks. By leveraging our high-fidelity V2X scene reconstructions and editable Gaussian representations, CRUISE enhances perception performance across ego-vehicle, infrastructure, and cooperative views. These findings highlight CRUISE’s potential as a versatile tool for advancing V2X perception through scalable, realistic data augmentation. In summary, this work makes the following key contributions:

\begin{figure*}
  \centering
  \includegraphics[width=1.0\textwidth]{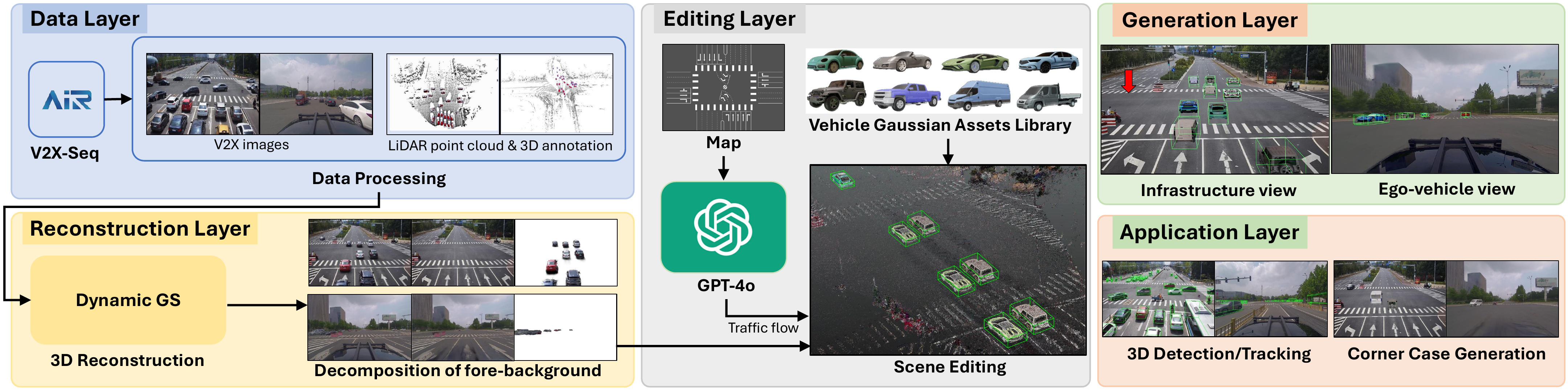}
  \vspace{-7mm}
  \caption{The workflow of CRUISE. The data layer processes the V2X-Seq dataset into suitable format for further reconstruction. The processed data is then used to decompose the fore-background scene based on dynamic GS modeling in the reconstruction layer. In the editing layer, the vector map and other information are fed into GPT-4o to generate possible traffic flows and edit the scene with vehicle Gaussian assets. The generation layer subsequently renders a new V2X dataset along with corresponding annotations. Finally, the newly generated dataset can be applied to 3D detection/tracking, corner case generation and further applications. The red arrow points to the position of the ego-car.}
  \label{fig:main}
  \vspace{-5mm}
\end{figure*}

\begin{itemize}
    \item We introduce CRUISE, the first GS-based simulation framework specifically designed for V2X driving scenarios, enabling high-fidelity reconstruction and multi-view synthesis.
    \item CRUISE supports high-fidelity scene reconstruction and flexible editing, allowing the generation of diverse and realistic V2X datasets from both ego-vehicle and infrastructure perspectives, facilitating better model training.
    \item Our data synthesis framework significantly improves 3D detection across ego-vehicle, infrastructure and cooperative views and cooperative 3D tracking.
    \item CRUISE further enables the generation of corner cases, enhancing dataset diversity and contributing to a data-driven closed-loop V2X driving system.
\end{itemize}

\section{Related Work}

\subsection{Autonomous Driving Simulation} 
Self-driving simulation engines, such as CARLA \cite{dosovitskiy2017carla} and AirSim \cite{shah2018airsim}, suffer from the costly manual effort to create virtual environments and the lack of realism in the generated data. In recent years, a lot of studies have been dedicated to building simulations from real-world autonomous driving data.
Former techniques \cite{manivasagam2020lidarsim, yang2020surfelgan, fang2020augmented,yang2023reconstructing, chen2021geosim, wang2022cadsim} usually concentrate on reconstructing through LiDAR or multi-view images, but fail to achieve high fidelity novel view synthesis (NVS).

Nowadays, the rapid advancement of 3D reconstruction techniques, including NeRF \cite{mildenhall2020nerf} and 3DGS \cite{kerbl3Dgaussians}, has attracted great attention in autonomous driving. For NeRF-based methods, Block-NeRF \cite{tancik2022block} and Mega-NeRF \cite{turki2022mega} reconstructed a large street scene by modeling segmented blocks. SUDS \cite{turki2023suds} and EmerNeRF \cite{yang2023emernerf} learn a scene decomposition of outdoor scenes. Moreover, other approaches \cite{Ost_2021_CVPR, KunduCVPR2022PNF, wu2023mars, yang2023unisim, ziyang2023snerf, tonderski2023neurad} utilize neural fields to model the scene as a combination of moving object networks and a background network. However, these methods suffer from high computational cost. 

For GS, PVG \cite{chen2023periodic} uses Periodic Vibration 3D Gaussians to model a dynamic urban scene. DrivingGaussian \cite{zhou2023drivinggaussian} introduces Composite Dynamic Gaussian Graphs and incremental 3D static Gaussians, while $S^3$Gaussian \cite{huang2024s3gaussian} differentiates between dynamic and static scenes in a self-supervised manner without the need for additional annotations. Although these GS-based methods are computationally efficient and high-fidelity, they do not achieve perfect reconstruction of both vehicles and streets. In this work, we utilize Street Gaussians~\cite{yan2024street}, which optimizes the tracked poses of dynamic Gaussians and introduces 4D spherical harmonics to vary vehicle appearances between frames, to achieve accurate deformation of vehicles and streets in V2X simulation.

\subsection{V2X Cooperative Perception}
V2X cooperative \cite{LiuWho2com:ICRA20,liu2020when2com,WangV2vnet:ECCV20,XuOPV2V:ICRA22,LiLearning:NeurIPS21,Li_2021_RAL,xu2022v2xvit,YuDAIRV2X:CVPR22,XuCoBEVT:CoRL22,HuWhere2comm:NeurIPS22,LuRobust:ICRA23,LiMultiRobot:CoRL22,GaoRegularized:RCS20,HuCollaboration:CVPR23,hu2024pragmatic} perception is an emerging application in V2X-aided systems that significantly enhances the autonomous driving perception module by allowing the exchange of complementary perceptual information. Several advanced platforms \cite{Li_2021_RAL,XuOPV2V:ICRA22,YuDAIRV2X:CVPR22,HuWhere2comm:NeurIPS22} have been developed to simulate cooperative perception scenarios, providing various perception annotations that support the development of these systems. In particular, cooperative perception systems have made substantial progress, with CoCa3D \cite{HuCollaboration:CVPR23} achieving near-complete perception capabilities. Furthermore, V2Xverse \cite{liu2024towards} has introduced an end-to-end cooperative driving system designed to facilitate V2X-based autonomous driving. 
In this work, we utilize the well-established cooperative perception benchmark, V2X-Seq \cite{yu2023v2x}, to evaluate the effectiveness of our simulation method. 

\section{Formulation \& Preliminaries}
3D-GS \cite{kerbl3Dgaussians} represents a 3D scene by a set of 3D Gaussians. The geometry of Gaussians is determined by $RSS^TR^T$, where $R \in \mathbb{R}^{3 \times 3}$ is the rotation matrix, and $S \in \mathbb{R}^{3 \times 3}$ is the scaling matrix.

To effectively model the surface by flattening the 3D Gaussians, we apply the scale loss proposed in \cite{chen2023neusg}. This loss minimizes the smallest component of the scaling factor $\mathbf{s} = (s_1, s_2, s_3)^\top \in \mathbb{R}^3$ for each Gaussian, driving it toward zero:
                    \begin{equation}
                        \mathcal{L}_{\text{scale}} =  \|\min(s_1,s_2,s_3)\|_1.
                    \end{equation}

To mitigate the needle-like artifacts during rendering, we propose setting $s_1$ as the longest scaling factor and $s_2$ as the second longest. Additionally, we apply a ratio loss to ensure that the Gaussian approximates a circular shape:
                    \begin{equation}
                        \mathcal{L}_{\text{ratio}} =  \max(1, s_1/s_2)-1.
                    \end{equation}

\section{Proposed Method}
CRUISE is designed to generate high-fidelity, virtually infinite V2X driving data through a dynamic GS-based reconstruction and rendering technique. The CRUISE pipeline is illustrated in Fig. \ref{fig:main}, and its hierarchical structure is divided into five distinct layers. The first data layer converts the V2X-Seq dataset into a format suitable for further reconstruction. The processed data is then used to decompose the foreground-background scene based on dynamic GS modeling in the reconstruction layer. In the editing layer, the vector map and other information are fed into a multimodal large language model to generate possible traffic flows and edit the scene with vehicle Gaussian assets. The generation layer subsequently renders a new V2X dataset and produces the corresponding annotations, facilitating downstream tasks. Finally, the newly generated dataset can be applied to 3D detection/tracking tasks, as well as other future applications.

\subsection{Data Processing and Scene Reconstructing}
The quality of the dataset significantly influences the outcomes of reconstruction and rendering processes. In this work, we use the meticulously collected real-world V2X dataset V2X-Seq \cite{yu2023v2x}, which contains rich RGB-images, point cloud from LiDAR and 3D annotations from both ego-vehicle and infrastructure views, as our data source.

Before data processing, we establish a GS-based reconstruction baseline, a point-based rendering approach that represents scene geometry with 3D Gaussians. GS methods have achieved state-of-the-art (SoTA) performance in 3D reconstruction and NVS, and their explicit representation is well-suited for our editing needs. We adopt Street Gaussians~\cite{yan2024street}, a top method for street scene reconstruction, due to its effective decomposition of objects and backgrounds, which facilitates data generation in our pipeline.

Next, we convert the data into the format required by Street Gaussians. Since the baseline method independently trains 4D objects and the 3D background using annotation boxes, the V2X dataset presents a challenge: the infrastructure view annotations include the box of the ego-vehicle.
When training on ego-vehicle views, the ego-car’s bounding box often encloses most of the visible scene, causing static background elements to be reconstructed as dynamic Gaussians. This issue undermines the proper decoupling of dynamic and static components. To address this issue, we remove the ego-box from the scene and introduce an ego-mask that covers ego-car regions in the ego-view, enabling more accurate separation of objects from the background.

For robust initialization, we fuse LiDAR point clouds from the ego-vehicle and infrastructure. To enhance object Gaussian density, we aggregate multi-frame point clouds using tracking boxes. Training incorporates LiDAR depth supervision from both views. Additionally, we adopt the appearance decoupling strategy from GOF \cite{Yu2024GOF}, which uses a lightweight convolutional neural network to model uneven illumination in real-world scenes, which reduces artifacts and improves high-fidelity rendering in data generation.

The loss function during GS training is set as follows:
\begin{equation}
    \label{eq:loss function}
    \begin{aligned}
        \mathcal{L} = \mathcal{L}_{\text{color}} + \lambda_1 \mathcal{L}_{\text{depth}} + \lambda_2 \mathcal{L}_{\text{normal}} + \lambda_3 \mathcal{L}_{\text{sky}}\\+ \lambda_4 \mathcal{L}_{\text{sem}} + \lambda_5 \mathcal{L}_{\text{scale}} + \lambda_6 \mathcal{L}_{\text{ratio}} + \lambda_7 \mathcal{L}_{\text{reg}}.
    \end{aligned}
\end{equation} 

In Eq. \eqref{eq:loss function}, $\mathcal{L}_{\text{color}}$ represents the reconstruction loss between the rendered and observed images with appearance decoupling following GOF \cite{Yu2024GOF}. 
$\mathcal{L}_{\text{depth}}$ and $\mathcal{L}_{\text{normal}}$ are the L1 loss between the rendered depth and normal, and the depth from LiDAR and normal generated by StableNormal \cite{ye2024stablenormal}.
$\mathcal{L}_{\text{sky}}$ is a binary cross-entropy loss for sky supervision with sky masks generated by Grounded SAM 2 \cite{ren2024grounded}. 
$\mathcal{L}_{\text{sem}}$ is a per-pixel softmax-cross-entropy loss between rendered semantic logits and input 2D semantic segmentation predictions \cite{li2022videoknet}. $\mathcal{L}_{\text{scale}}$ and $\mathcal{L}_{\text{ratio}}$ help constrain the geometry of Gaussians to a flattened circular shape to improve surface reconstruction. Finally, $\mathcal{L}_{\text{reg}}$ is a regularization term used to remove floaters and enhance decomposition effects. 
\begin{figure}
  \centering
  \includegraphics[width=0.48\textwidth]{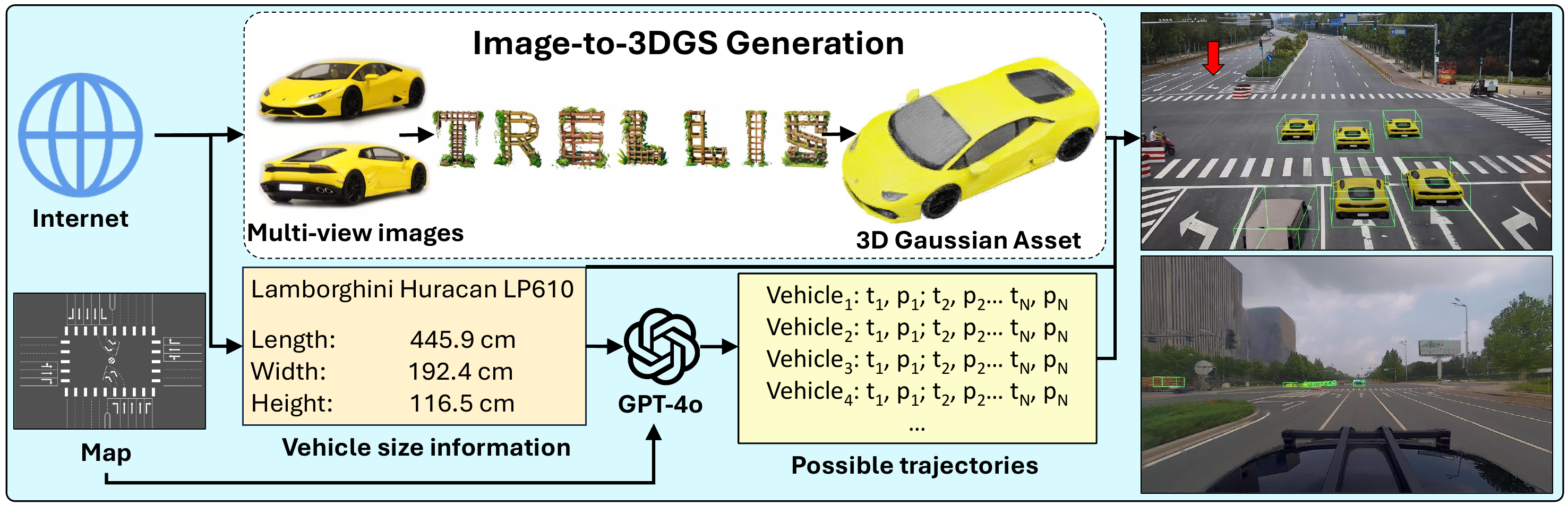}
  \vspace{-7mm}
  \caption{The pipeline of scene editing. By collecting basic vehicle information from the Internet, multi-view images are processed using TRELLIS to generate 3D Gaussian vehicle assets. Simultaneously, the vehicle size, vector map, and ego-car trajectory are provided as inputs to GPT-4o, which outputs possible trajectories in the form of frame indices $\mathbf{t}_i$ and corresponding poses $\mathbf{p}_i$). Finally, the vehicle asset, box size, and generated trajectories are used to place the vehicle within the scene to producing new V2X data. The red arrow indicates the position of the ego-car.}
  \label{fig:edit}
  \vspace{-5mm}
\end{figure}

\begin{figure*}
\begin{center}
\begin{tabular}{{@{}c@{\hspace{2pt}}c@{\hspace{2pt}}c@{\hspace{2pt}}c@{\hspace{2pt}}c@{}}}
\includegraphics[width=4.35cm]{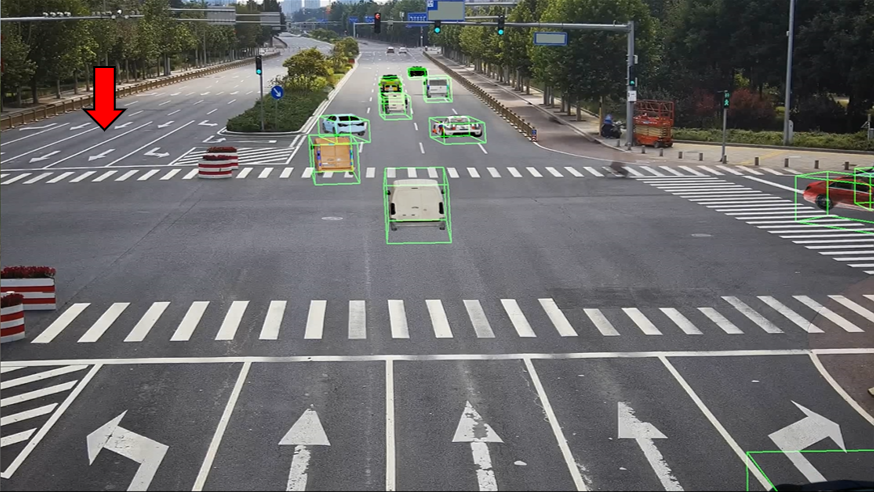}&
\includegraphics[width=4.35cm]{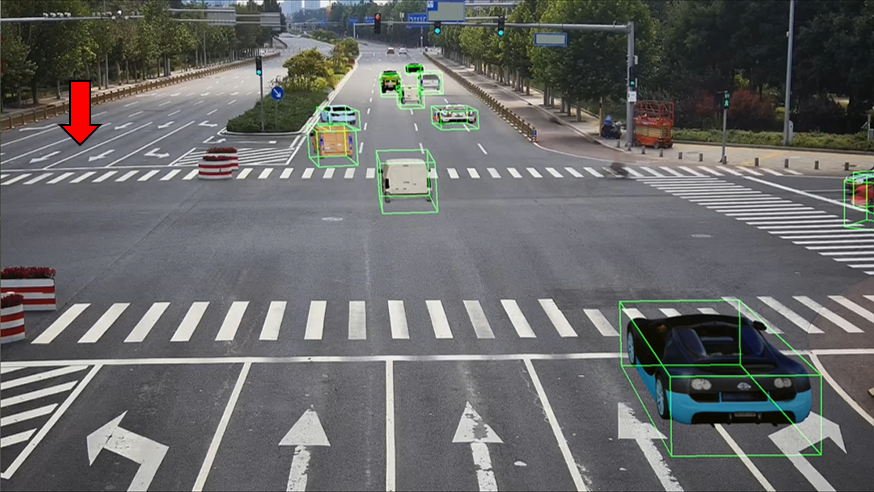}&
\includegraphics[width=4.35cm]{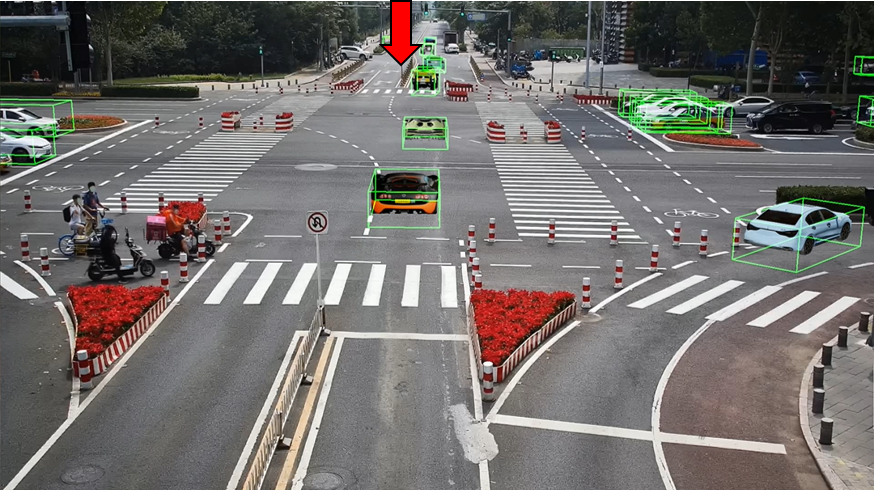}&
\includegraphics[width=4.35cm]{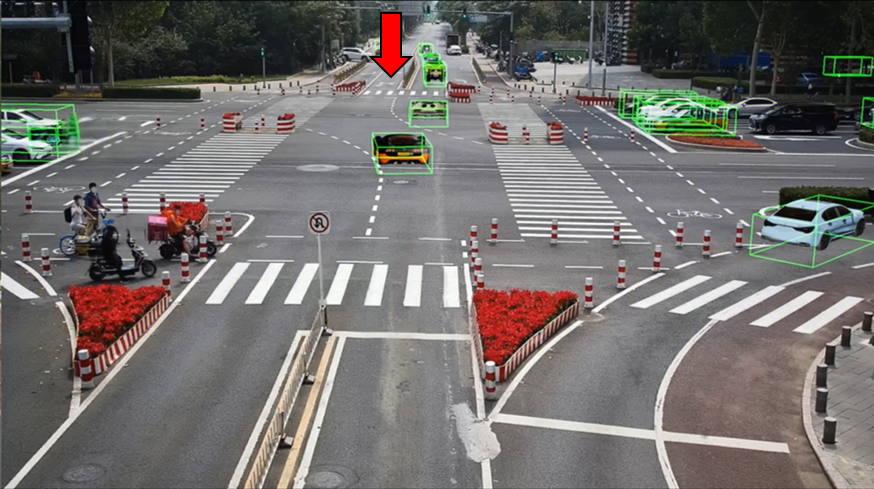}\\
\end{tabular}
\begin{tabular}{{@{}c@{\hspace{2pt}}c@{\hspace{2pt}}c@{\hspace{2pt}}c@{\hspace{2pt}}c@{}}}
\includegraphics[width=4.35cm]{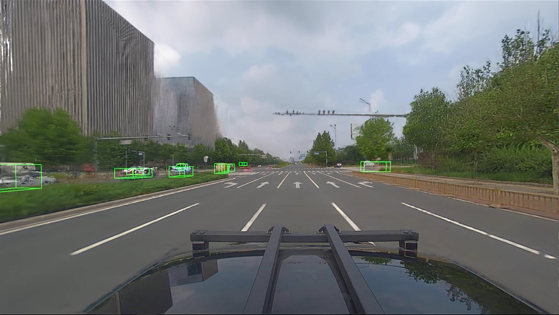}&
\includegraphics[width=4.35cm]{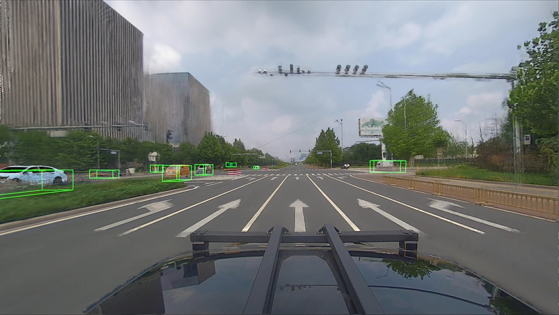}&
\includegraphics[width=4.35cm]{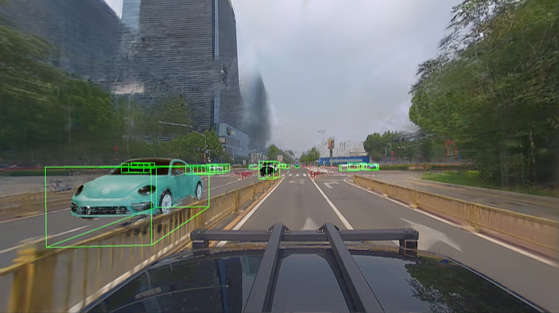}&
\includegraphics[width=4.35cm]{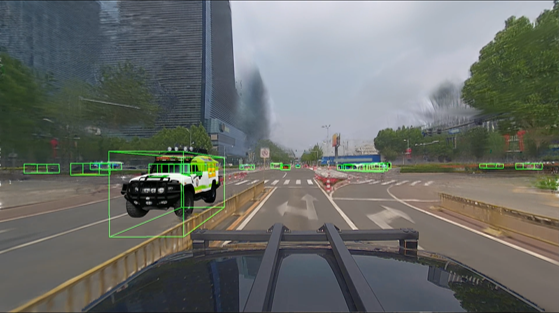}\\
\end{tabular}
\end{center}
\vspace{-4mm}
\caption{Data generated by CRUISE. Up-to-down: Infrastructure views; Corresponding Ego-vehicle views. The red arrow points to the position of the ego-car.}
\label{fig:gen}
\vspace{-5mm}
\end{figure*}

\subsection{Scene Editing and Data Generating}
After reconstruction, vehicles and streets can be distinctly decomposed as depicted in Fig. \ref{fig:main}, which facilitates flexible scene editing.

For editing, we design a generation-based pipeline shown in Fig. \ref{fig:edit}. First, we collect multi-view vehicle images along with their sizes from the Internet. These images are then fed into TRELLIS \cite{xiang2024structured}, a SoTA 3D generation method, to produce high-quality 3D Gaussian vehicle assets with accurate geometry and appearance. Simultaneously, the vehicle size, vector map, and ego-car trajectory are provided to GPT-4o to generate possible trajectories for scene insertion.
Specifically, since all scenes are intersections, we first simplify the vector map. Based on the LiDAR pose of the infrastructure, we select the nearest lane labeled as CITY\_DRIVING and marked with the is\_intersection flag. We then iteratively add adjacent lanes using a breadth-first search approach until all relevant lanes in the sequence are included, as verified by visualizing the centerlines of the lanes. Next, we provide the starting positions and directions of each lane's centerline as input to GPT-4o. In addition, we provide the ego-car trajectory sampled every 10 frames as a sequence:
$\mathbf{t}_1$, $\mathbf{p}_1$; $\mathbf{t}_2$, $\mathbf{p}_2$; …,
where $\mathbf{t} \in {0, 10, 20, \dots, N}$ denotes the frame index and $\mathbf{p} \in \mathbb{R}^3$ represents the position. GPT-4o then leverages its multimodal reasoning capabilities to generate plausible trajectories for each additional vehicle, in the same format as the ego-car’s trajectory. We apply interpolation to estimate positions at every frame and, together with the vehicle size, construct tracking boxes for inserting Gaussian vehicles into the scene.

After editing, we render new V2X images from both ego-vehicle and infrastructure perspectives for every frame. For the ego-vehicle view, we paste back the ego-car region to more faithfully reproduce the original data.
In parallel, we generate 3D annotation boxes for all vehicles, resulting in a fully labeled, synthetic V2X dataset. Rendered images with annotation boxes are shown in Fig. \ref{fig:gen}.

\subsection{Downstream Application}
The downstream applications of CRUISE include several aspects, including 3D detection/tracking, corner case generation, and closed-loop training and testing for V2X tasks. 

As end-to-end autonomous driving advances, V2X communication is becoming increasingly essential for safety and full autonomy. However, most research prioritizes algorithmic improvements, overlooking the critical role of large, diverse datasets for effective training.

CRUISE reconstructs real-world scenes and generates extensive V2X datasets, including out-of-distribution (OOD) cases, enabling the training of more complex models with enhanced accuracy and generalization. Additionally, CRUISE facilitates corner case generation, helping autonomous systems better handle challenging driving scenarios. With its robust data synthesis capabilities, CRUISE has the potential to establish a benchmark for closed-loop training and evaluation in future V2X research.

In this work, we primarily evaluate our generated data on V2X tasks such as 3D object detection and tracking, demonstrating the effectiveness of the CRUISE framework.

\section{Experiments}

\subsection{Dataset}
\noindent\textbf{V2X-Seq.}
The V2X-Seq dataset \cite{yu2023v2x} is a benchmark built on the DAIR-V2X-C dataset \cite{YuDAIRV2X:CVPR22} for vehicle-infrastructure cooperative perception in autonomous driving. It contains real-world data captured at 10 Hz across 95 traffic scenes at 6 intersections, with each scene lasting 10 to 20 seconds. The dataset includes high-frequency recordings from vehicles and infrastructure units, each equipped with LiDAR and a camera, offering a multi-modal view of traffic dynamics. It also provides 3D tracking annotations for each object of interest in each sequence, with unique tracking IDs shared by the same objects in each sequence, with a vector map.

\subsection{Implementation Details}

All experiments are conducted on an NVIDIA A800 GPU. To assess the efficiency of our framework, we selected six sequences from 4 intersections of the V2X-Seq dataset to reconstruct the scene, generate new data, and evaluate the effectiveness of the generated data in downstream tasks.

For reconstruction, we train Street Gaussians for 50,000 iterations. For 3D detection, we choose the SoTA methods on the benchmark: MonoLSS \cite{li2024monolss} for vehicle-only views, Bevheight \cite{yang2023bevheight} for infrastructure-only views, and ImVoxelNet \cite{rukhovich2022imvoxelnet} for cooperative views. 
ImvoxelNet can also perform cooperative 3D tracking.  
MonoLss is trained for 150 epochs, Bevheight for 100 epochs, and ImvoxelNet for 24 epochs.

\subsection{Comparison on Reconstruction}
Before the editing and generating, we compare the reconstruction results of different methods. We test on GS-based baselines, 3D-GS \cite{kerbl3Dgaussians}, PVG \cite{chen2023periodic}, $\textit{S}^3$Gaussian \cite{huang2024s3gaussian}, HUGS~\cite{zhou2024hugs}, and Street Gaussians \cite{yan2024street} on the same 6 sequences of V2X-Seq. We report the average \textbf{PSNR} (peak signal-to-noise ratio), \textbf{SSIM} (structural similarity index)
and \textbf{LPIPS} (learned perceptual image patch similarity).

From Tab \ref{tab:recon}, the Street Gaussians we used synthetically outperforms other methods in the rendering of scene reconstruction. Additionally, by utilizing the object box to distinguish between objects and the background during training, it facilitates subsequent scene editing tasks.

\subsection{Ablation Studies on Reconstruction Modules}
Table \ref{tab:ab} presents the quantitative results of our designed modules on V2X-Seq.

\noindent\textbf{Effectiveness of the Loss Function.} 
The results presented in Table \ref{tab:ab} demonstrate that the inclusion of normal loss and geometric losses ($\mathcal{L}_{\text{scale}}$ and $\mathcal{L}_{\text{ratio}}$) helps to improve the geometry modeling of the scene, leading to high-fidelity reconstruction. 

\noindent\textbf{Effectiveness of the Ego-mask.}
Table \ref{tab:ab} clearly shows that the absence of the ego-mask leads to ambiguity in distinguishing the ego-car parts from the background in the ego views, resulting in poor reconstruction and data generation performance.

\noindent\textbf{Effectiveness of the appearance decoupling.}
From Table \ref{tab:ab}, including appearance decoupling improves the reconstruction results, since it facilitates Gaussians to learn consistent geometry and colors, rather than compensating for
view-dependent appearance.

\begin{table}[]
  \centering
  \fontsize{9pt}{10pt}\selectfont
  \caption{Results of reconstruction on V2X-Seq. $\textit{S}^3$-GS is $\textit{S}^3$Gaussian, Street-GS is Street Gaussians.}
    \begin{tabular}{lc|ccc}
    \toprule
    Method & Box & PSNR $\uparrow$ & SSIM $\uparrow$ & LPIPS $\downarrow$ \\
    \midrule
    3D-GS \cite{kerbl3Dgaussians} & & 24.79 & 0.902 & 0.158 \\
    PVG \cite{chen2023periodic} & & \cellcolor{tabfirst}\textbf{28.51} & 0.924 & 0.115 \\
    $\textit{S}^3$-GS \cite{huang2024s3gaussian} & & \cellcolor{tabthird}27.54 & \cellcolor{tabsecond}0.931 & \cellcolor{tabsecond}0.097 \\
    HUGS \cite{zhou2024hugs} & \checkmark & 27.23 & \cellcolor{tabthird}0.925 & \cellcolor{tabthird}0.101 \\
    \midrule
    \textbf{Street-GS(Ours) \cite{yan2024street}} & \checkmark & \cellcolor{tabsecond}27.97 & \cellcolor{tabfirst}\textbf{0.940} & \cellcolor{tabfirst}\textbf{0.095} \\
    \bottomrule
    \end{tabular}%
  \label{tab:recon}%
\end{table}%

\begin{table}[]
  \centering
  \fontsize{9pt}{10pt}\selectfont
  \caption{Ablation studies on reconstruction modules}
    \begin{tabular}{l|ccc}
    \toprule
    \multicolumn{1}{c|}{Methods} & PSNR $\uparrow$ & SSIM $\uparrow$ & LPIPS $\downarrow$ \\
    \midrule
    Ours w/o Normal loss & 27.45 & 0.933 & 0.103\\
    Ours w/o Geo. loss & 27.37 & 0.935 & 0.100\\
    Ours w/o Ego-mask & 24.44 & 0.910 & 0.125\\
    Ours w/o Appearance & 26.89 & 0.929 & 0.117\\
    Complete method & \textbf{27.97} & \textbf{0.940} & \textbf{0.095} \\
    \bottomrule
    \end{tabular}%
  \label{tab:ab}%
  \vspace{-4mm}
\end{table}%

\begin{figure*}
\begin{center}
\begin{tabular}{{@{}c@{\hspace{2pt}}c@{\hspace{2pt}}c@{\hspace{2pt}}c@{\hspace{2pt}}c@{}}}
\includegraphics[width=4.35cm]{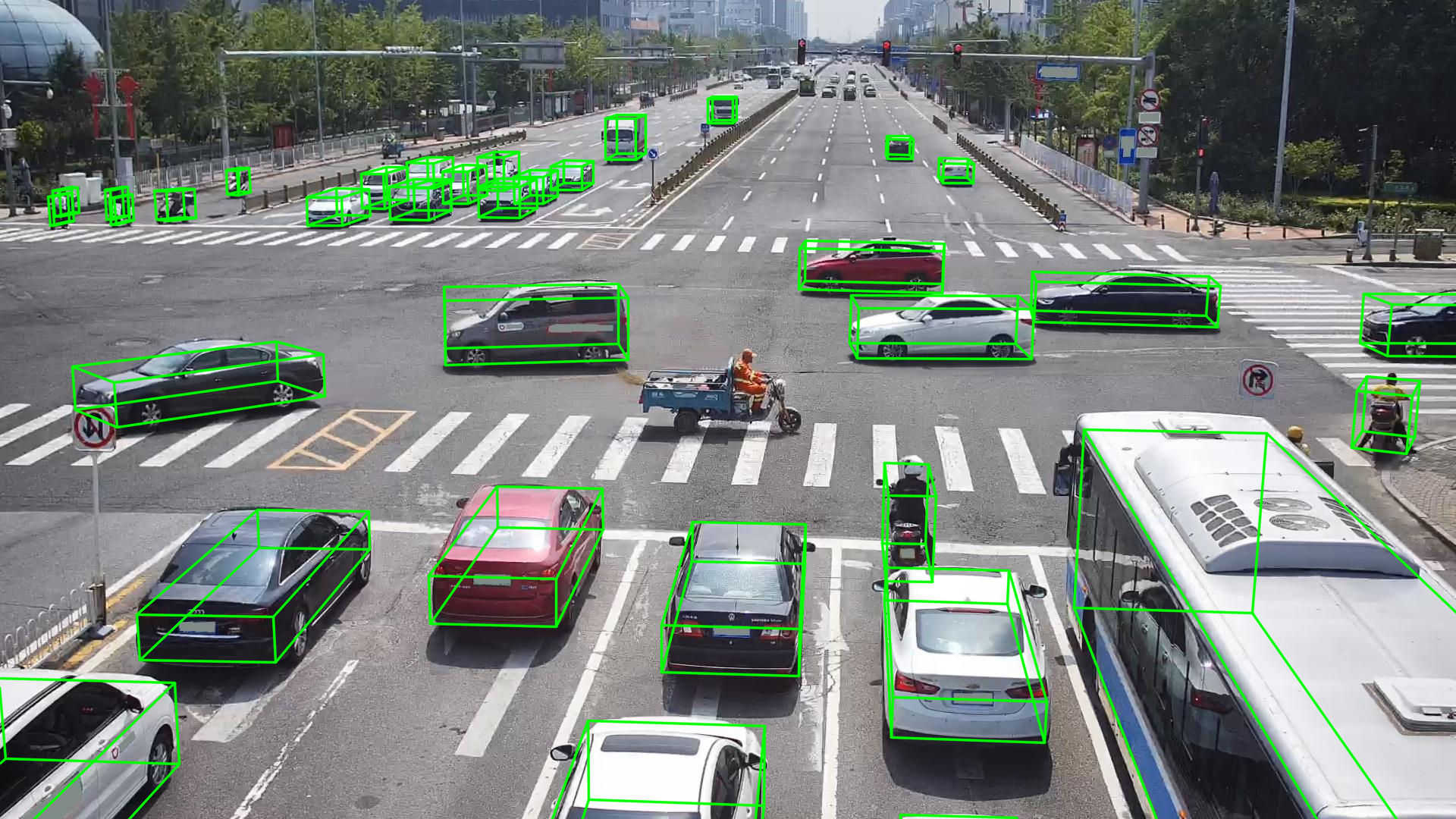}&
\includegraphics[width=4.35cm]{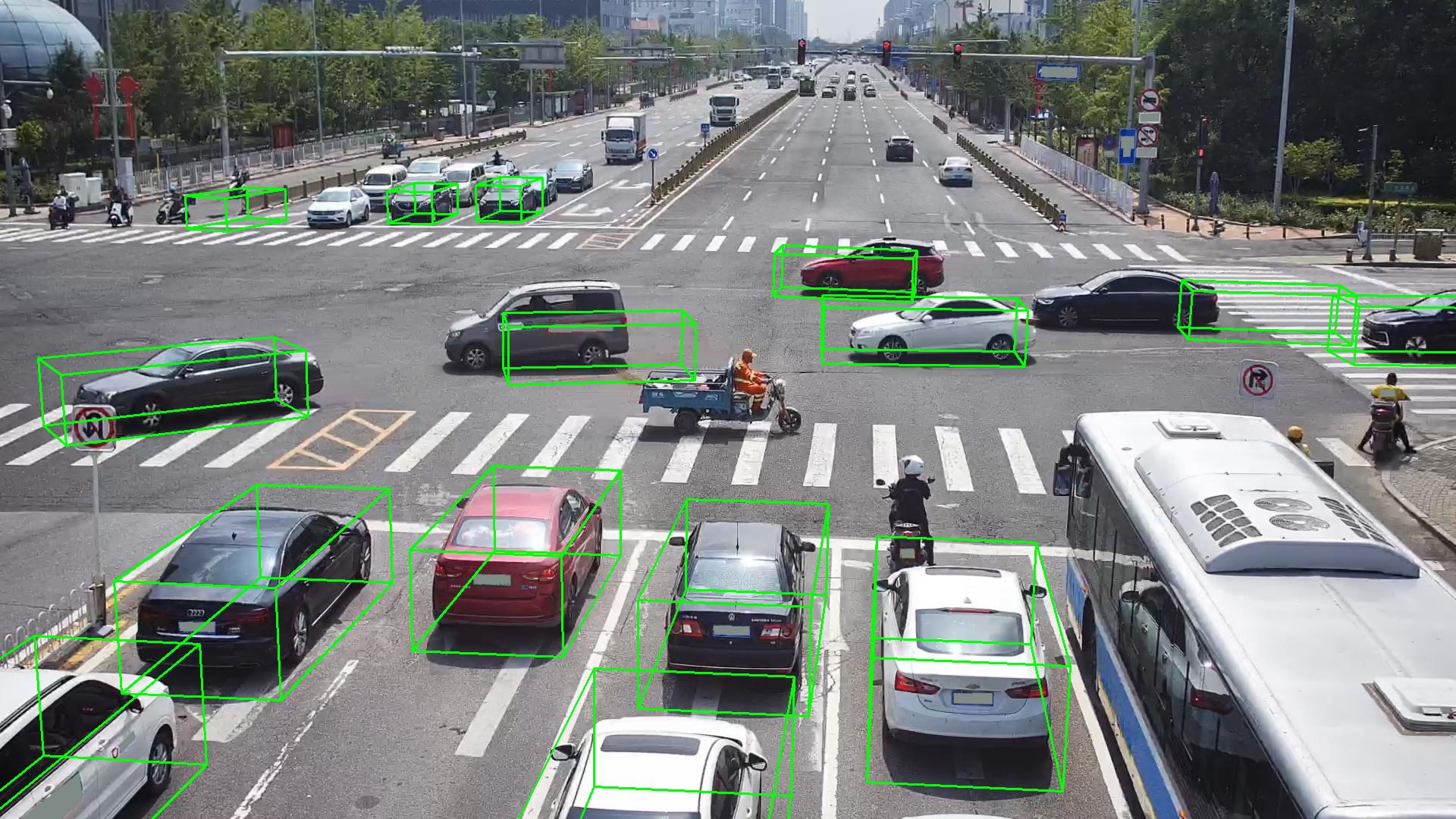}&
\includegraphics[width=4.35cm]{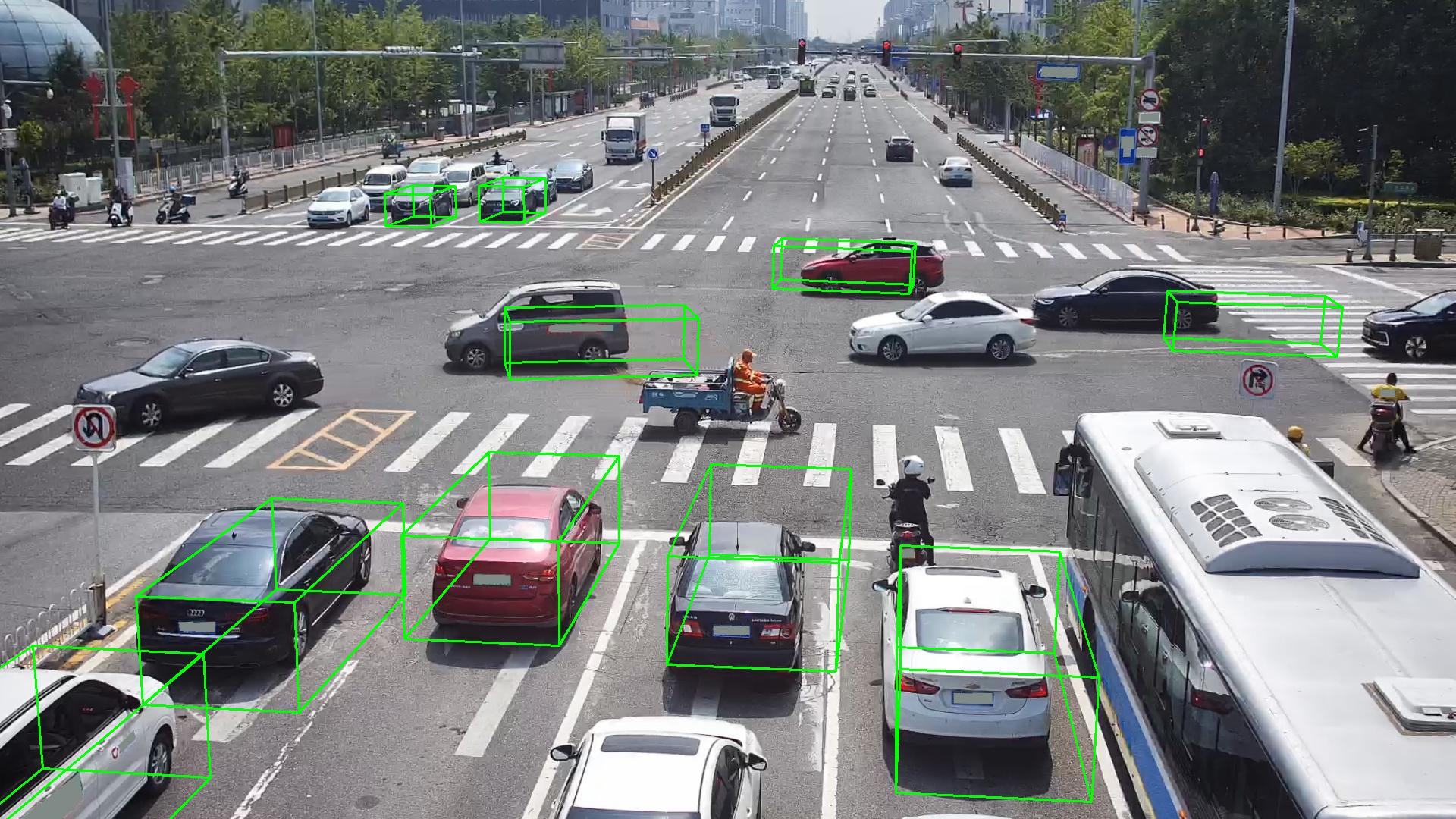}&
\includegraphics[width=4.35cm]{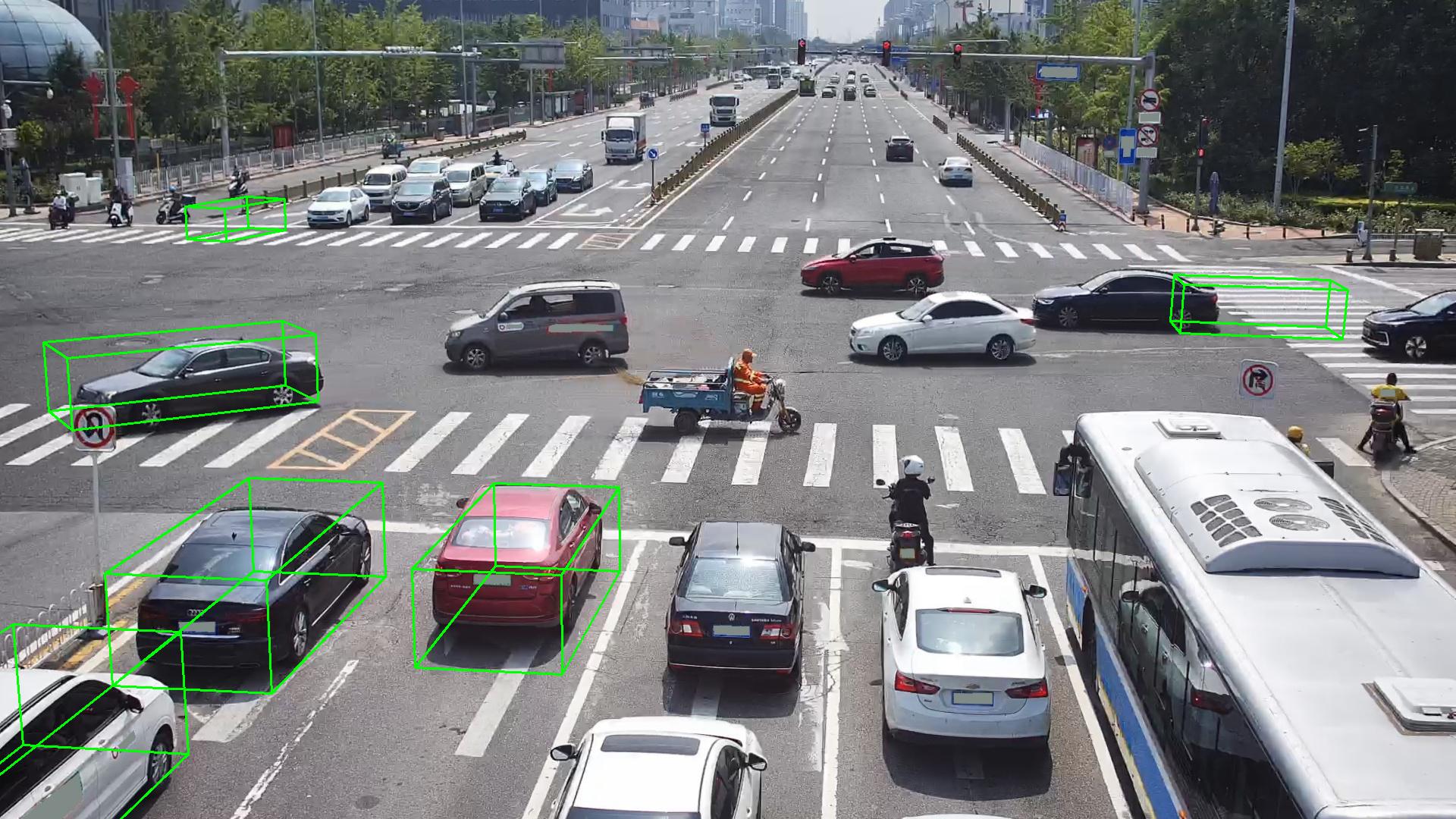}\\
\end{tabular}
\begin{tabular}{{@{}c@{\hspace{2pt}}c@{\hspace{2pt}}c@{\hspace{2pt}}c@{\hspace{2pt}}c@{}}}
\includegraphics[width=4.35cm]{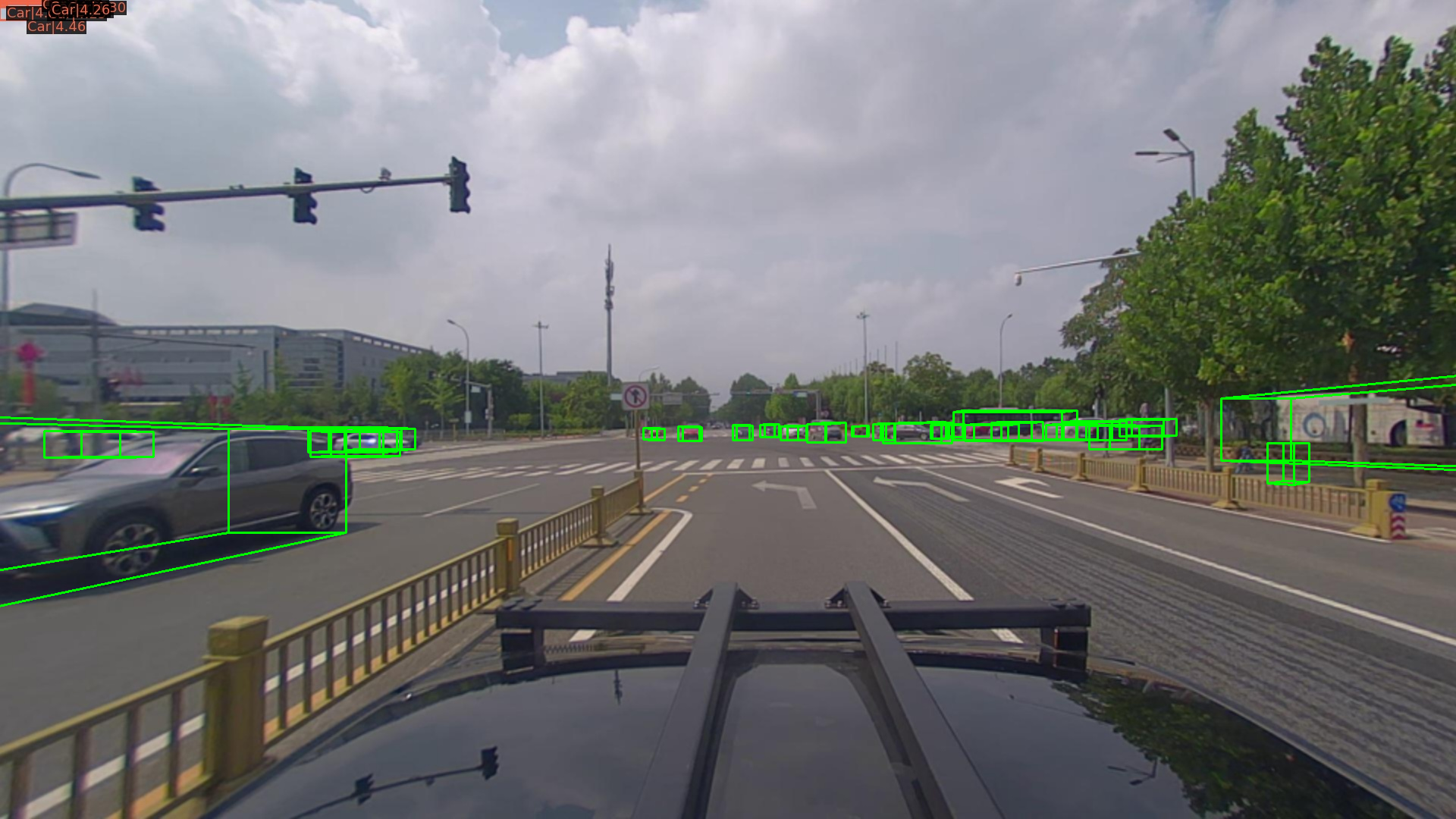}&
\includegraphics[width=4.35cm]{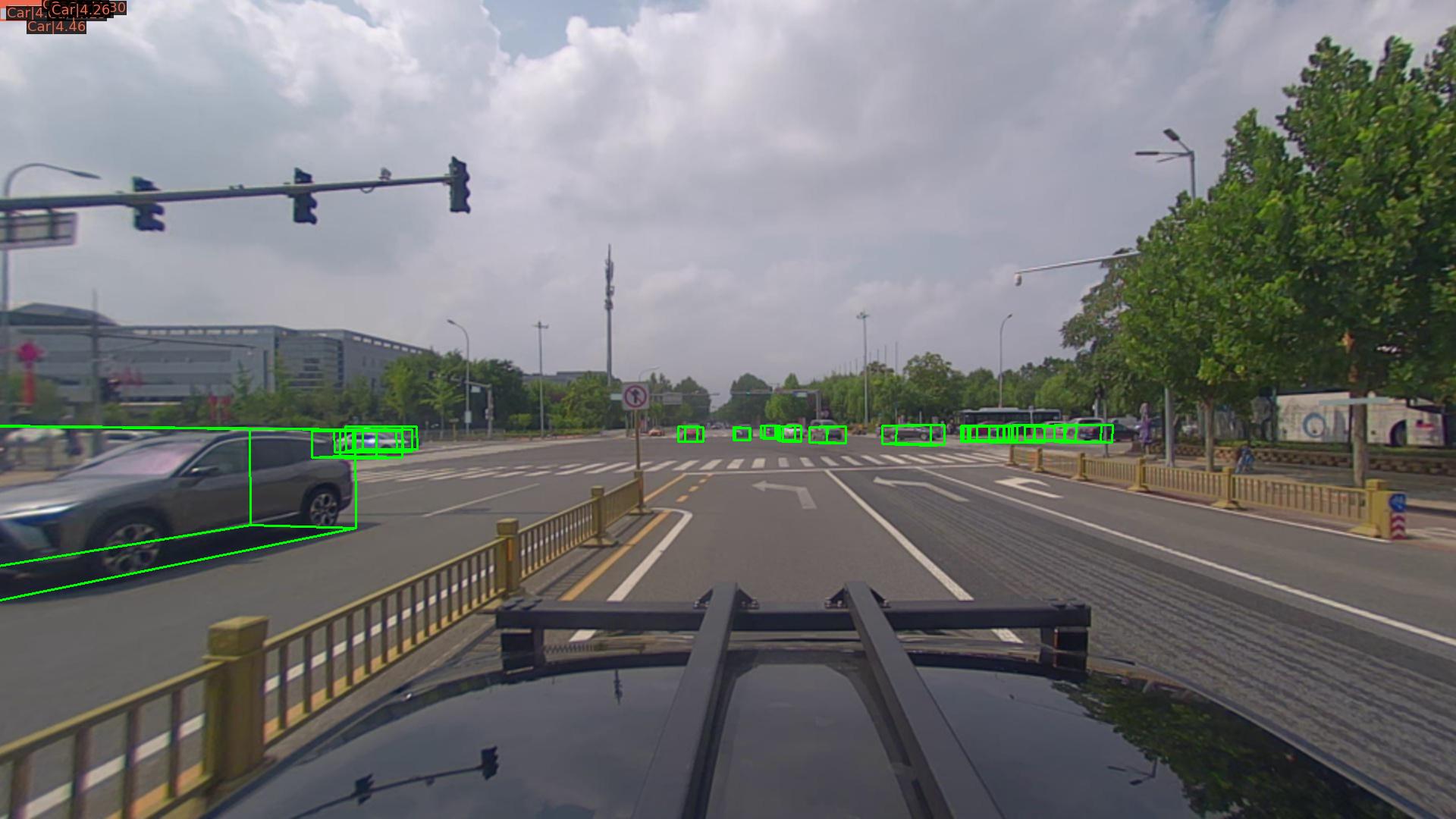}&
\includegraphics[width=4.35cm]{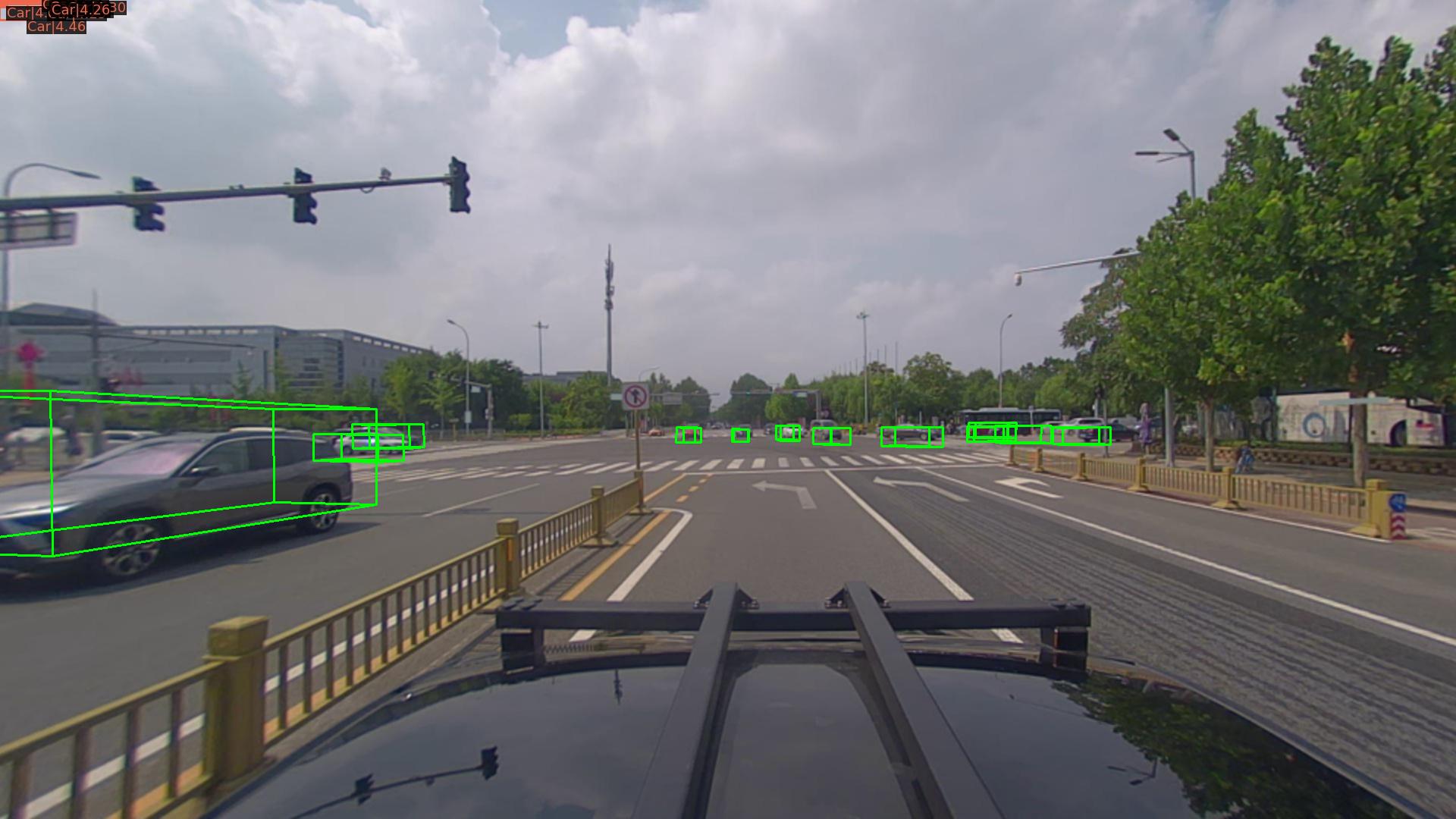}&
\includegraphics[width=4.35cm]{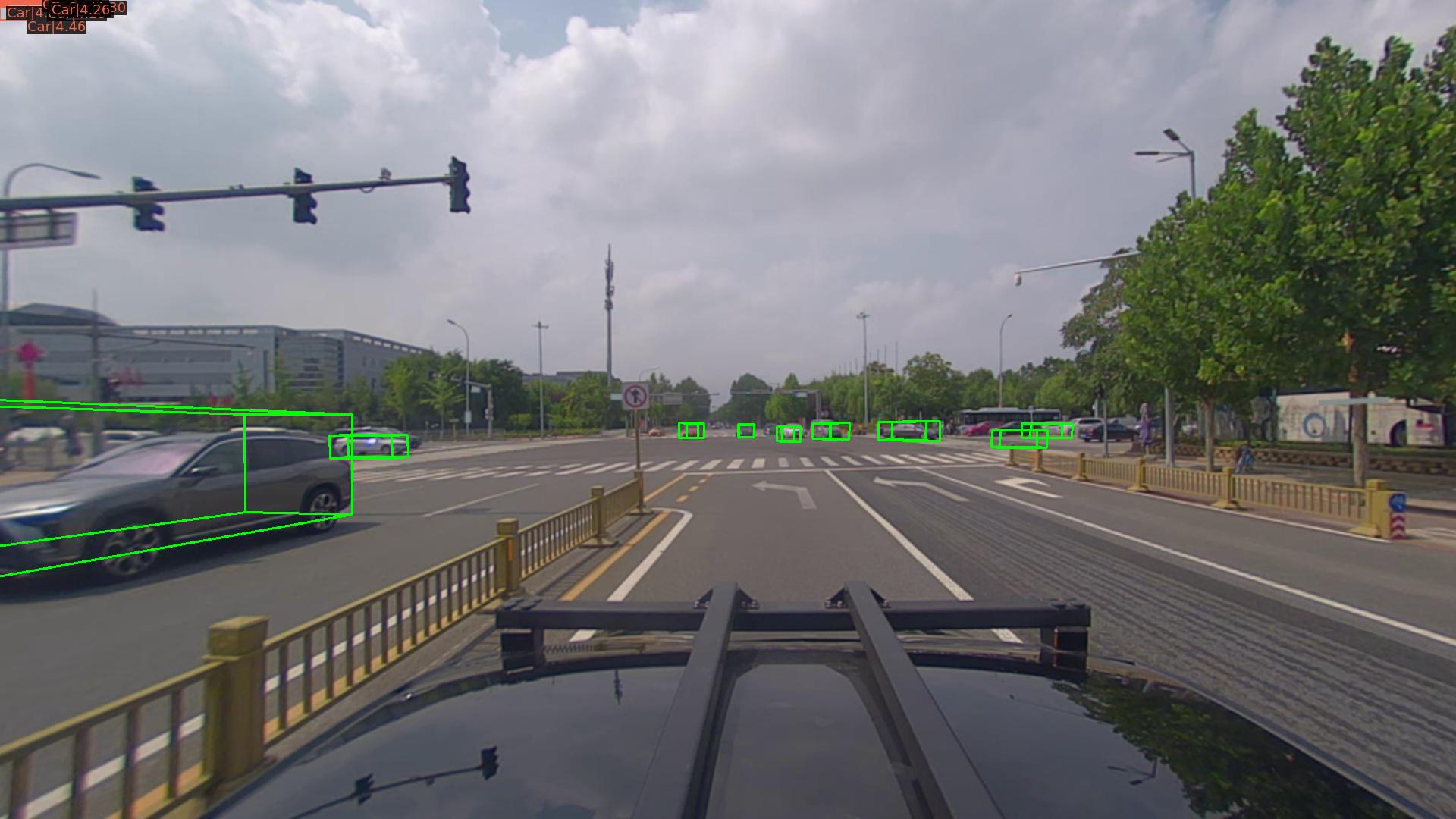}\\
\end{tabular}
\end{center}
\vspace{-4mm}
\caption{Qualitative results of cooperative 3D detection on the original V2X-Seq dataset. The upper row shows infrastructure views, while the lower row shows corresponding ego-vehicle views. From left to right: Ground truth; Real + Gen; Gen; Real. The results demonstrate improved detection accuracy and vehicle recognition with the use of augmented data for training.}
\vspace{-5mm}
\label{fig:detection}
\end{figure*}

\subsection{Results on 3D Detection/Tracking}
\begin{table}[]
  \centering
  \fontsize{8.5pt}{9.5pt}\selectfont
  \caption{Quantitative results of the vehicle view 3D detection on V2X-Seq}
    \begin{tabular}{l|ccc|ccc}
    \toprule
    \multicolumn{1}{c|}{\multirow{2}[4]{*}{Train data}} & \multicolumn{3}{c|}{$AP_{3D}(IOU=0.7)\uparrow$} & \multicolumn{3}{c}{{$AP_{BEV}(IOU=0.7)\uparrow$}} \\
\cmidrule{2-7}           & Easy & Mod. & Hard & Easy & Mod. & Hard \\
    \midrule
    Real & \cellcolor{tabsecond}56.84 & \cellcolor{tabsecond}37.69 & \cellcolor{tabsecond}34.14 & \cellcolor{tabsecond}66.81 & \cellcolor{tabthird}46.62 & \cellcolor{tabthird}44.21 \\
    Gen & \cellcolor{tabthird}54.41 & \cellcolor{tabthird}36.77 & \cellcolor{tabthird}34.02 & \cellcolor{tabthird}65.44 & \cellcolor{tabsecond}47.13 & \cellcolor{tabsecond}45.33 \\
    Real + Gen &\bfseries \cellcolor{tabfirst}61.35 &\bfseries \cellcolor{tabfirst}41.05 &\bfseries \cellcolor{tabfirst}38.71 &\bfseries \cellcolor{tabfirst}67.38 &\bfseries \cellcolor{tabfirst}49.72 &\bfseries \cellcolor{tabfirst}47.24 \\
    \bottomrule
    \end{tabular}%
  \label{tab:3}%
\end{table}%

\begin{table}[]
  \centering
  \fontsize{8.5pt}{9.5pt}\selectfont
  \caption{Quantitative results of the infrastructure view 3D detection on V2X-Seq}
    \begin{tabular}{l|ccc|ccc}
    \toprule
    \multicolumn{1}{c|}{\multirow{2}[4]{*}{Train data}} & \multicolumn{3}{c|}{$AP_{3D}(IOU=0.7)\uparrow$} & \multicolumn{3}{c}{{$AP_{BEV}(IOU=0.7)\uparrow$}} \\
\cmidrule{2-7}           & Easy & Mod. & Hard & Easy & Mod. & Hard \\
    \midrule
    Real & \cellcolor{tabthird}60.03 & \cellcolor{tabthird}50.16 & \cellcolor{tabthird}50.13 & \cellcolor{tabthird}74.86 & \cellcolor{tabsecond}62.54 & \cellcolor{tabsecond}62.53 \\
    Gen & \cellcolor{tabsecond}63.72 & \cellcolor{tabsecond}52.22 & \cellcolor{tabsecond}52.19 & \cellcolor{tabsecond}75.78 & \cellcolor{tabthird}61.65 & \cellcolor{tabthird}61.64 \\
    Real + Gen &\bfseries \cellcolor{tabfirst}64.10 &\bfseries \cellcolor{tabfirst}54.26 &\bfseries \cellcolor{tabfirst}52.40 &\bfseries \cellcolor{tabfirst}76.35 &\bfseries \cellcolor{tabfirst}64.19 &\bfseries \cellcolor{tabfirst}64.17 \\
    \bottomrule
    \end{tabular}%
  \label{tab:4}%
\end{table}%

\begin{table}[]
  \centering
  \fontsize{9pt}{10pt}\selectfont
  \caption{Quantitative results of the cooperative view 3D detection/tracking on V2X-Seq}
    \begin{tabular}{l|cc|cc}
    \toprule
    \multicolumn{1}{c|}{\multirow{3}[2]{*}{Train data}} & \multicolumn{2}{c|}{3D Detection} & \multicolumn{2}{c}{3D Tracking} \\
          & {$AP_{3D}\uparrow$} & {$AP_{BEV}\uparrow$} & \multirow{2}[1]{*}{MOTA$\uparrow$} & \multirow{2}[1]{*}{MOTP$\uparrow$} \\
          & (IOU=0.5) & (IOU=0.5) &       &  \\
    \midrule
    Real  & \cellcolor{tabthird}14.79 & \cellcolor{tabthird}19.75 & \cellcolor{tabthird}21.83 & \cellcolor{tabthird}56.65 \\
    Gen & \cellcolor{tabsecond}14.99 & \cellcolor{tabsecond}20.09 & \cellcolor{tabsecond}25.03 & \cellcolor{tabsecond}57.29 \\
    Real + Gen &\bfseries \cellcolor{tabfirst}15.91 &\bfseries \cellcolor{tabfirst}20.74 &\bfseries \cellcolor{tabfirst}25.52 &\bfseries \cellcolor{tabfirst}58.15 \\
    \bottomrule
    \end{tabular}%
  \label{tab:5}%
  \vspace{-4mm}
\end{table}%

For downstream tasks, we construct a test set with 8 sequences from 4 intersections in V2X-Seq, excluding those used for data generation. For the training set, we generated 1000 frames on the reconstructed scenes and selected the same number of frames from the real-world sequences used for reconstruction as a fair comparison.

\noindent\textbf{Results on Single View 3D Detection.} 
As shown in Table \ref{tab:3}, the generated data achieves results comparable to real data in both ends of the 3D detection tasks, demonstrating that the generated data can match or even surpass real data in terms of authenticity and label accuracy. To further assess its diversity, we incorporated generated data as a form of data augmentation, training it alongside real data for half the training iterations. Table \ref{tab:3} shows that this hybrid training strategy significantly improves detection model performance, indicating that the generated data provides high-quality OOD samples that enhance model generalization. These results indicate that generated data can serve as a valuable supplement in practical applications, offering robust and reliable training samples for single view 3D detection tasks.

\noindent\textbf{Results on Cooperative View 3D Detection and Tracking.}
The results in Table \ref{tab:5} indicate that in the vehicle-infrastructure cooperative task, models trained on generated data achieve performance comparable to or exceeding those trained on real data, particularly in 3D tracking, where a significant improvement is observed. This advantage may stem from the fact that the generated data is largely free from missed annotations and is labeled with precise bounding boxes following the vehicle trajectories. As a result, tracking annotations are smoother and more accurate, enhancing model performance. Additionally, training with a combination of real and generated data for half of the training steps also exhibits similar performance improvements as single views, demonstrating that generated data can serve as an effective data augmentation strategy to enhance model robustness. The qualitative results in Fig. \ref{fig:detection} show that models trained solely on real data struggle with OOD scenarios, leading to poor detection performance. These findings suggest that CRUISE-generated data introduces more diverse sequences, enhancing the model’s generalization capability. These results showcase CRUISE's potential as a comprehensive V2X autonomous driving simulation platform.

\section{Discussion}
\begin{figure}
\begin{center}
\begin{tabular}{{@{}c@{}}}
\includegraphics[width=0.238\textwidth]{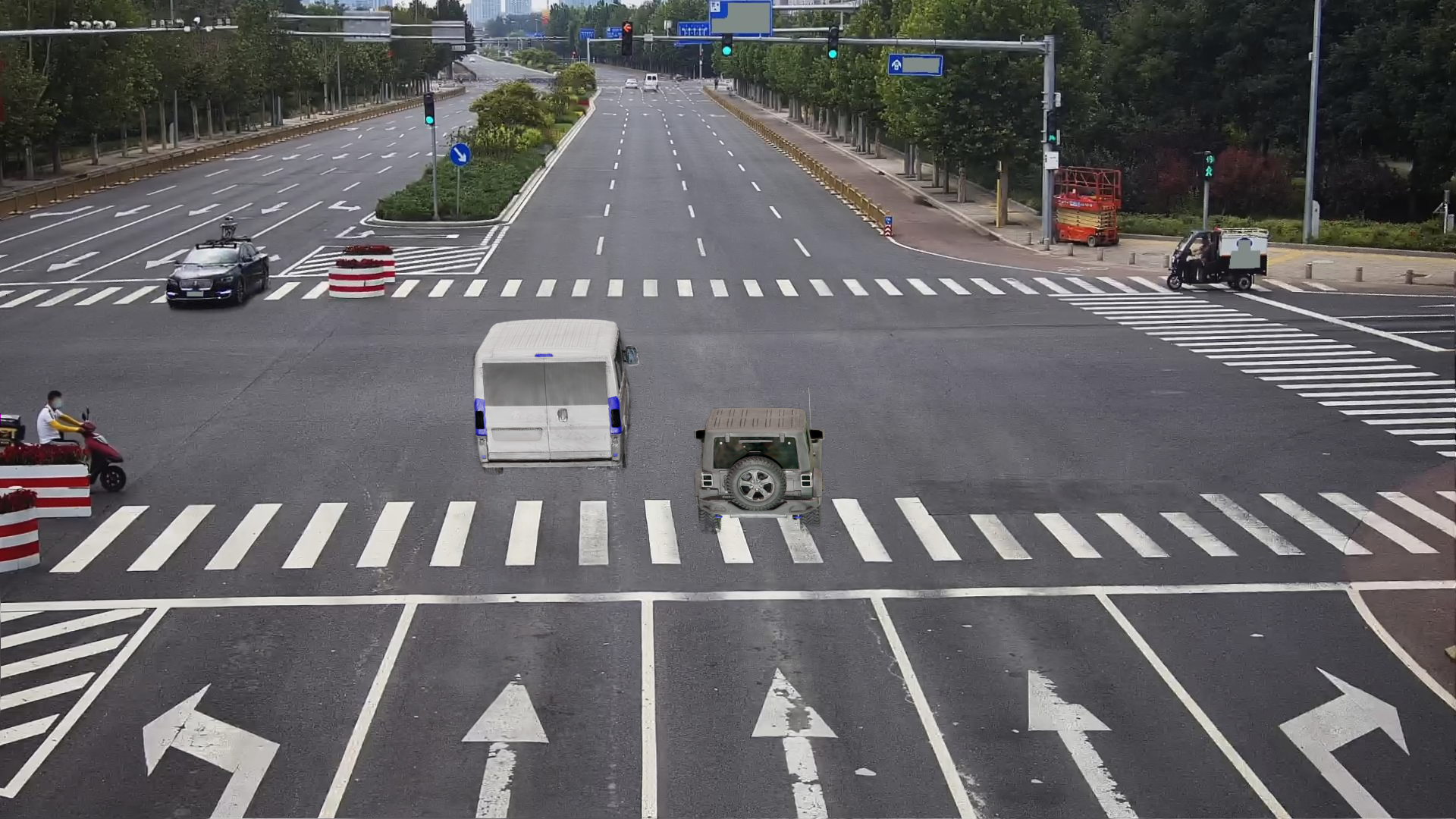}\\(a)\\
\end{tabular}
\begin{tabular}{{@{}c@{}}}
\includegraphics[width=0.238\textwidth]{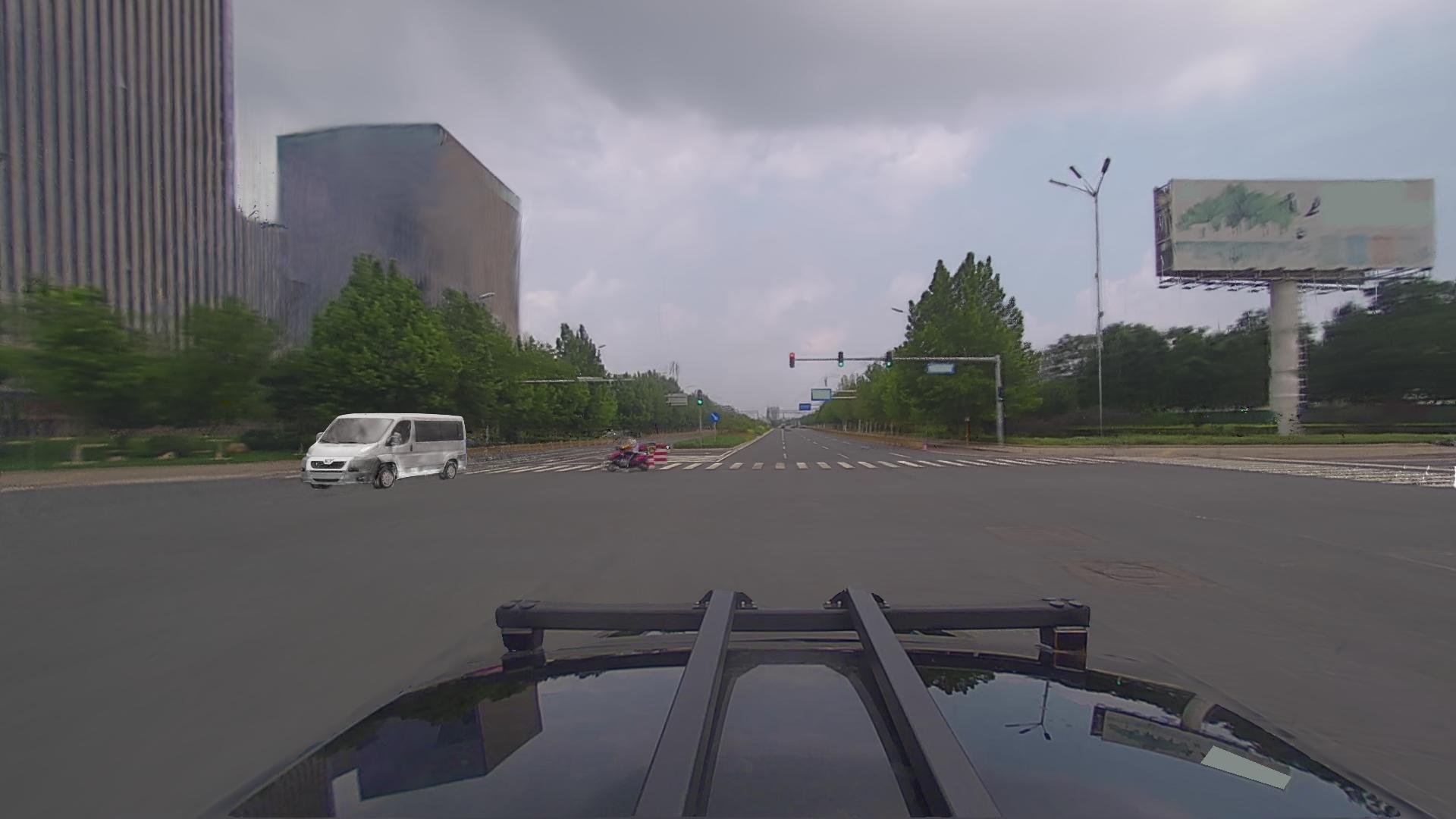}\\(b)\\
\end{tabular}
\end{center}
\vspace{-3mm}
\caption{The visualization of the corner case: A demonstration of the vehicle-side occlusion scenario. (a) Infrastructure view; (b) Ego-vehicle view.}
\vspace{-4mm}
\label{fig:cc}
\end{figure}

\subsection{Corner Case Generation.}
CRUISE not only enhances 3D detection and tracking but also enables the generation of critical corner cases in driving scenarios. As shown in Fig. \ref{fig:cc}, two vehicles are visible from the infrastructure view (a), but in the ego-vehicle view (b), the green jeep is occluded by the white van. If the ego-vehicle turns left, V2X communication can relay this hidden information, enabling more informed decision-making for autonomous driving.
Simulating such corner cases is essential for V2X-enabled autonomous driving, providing a cost-effective and scalable method to improve system safety and robustness in complex real-world scenarios.

\subsection{Synchronized Generation.}
Many real-world V2X datasets suffer from timestamp misalignment between roadside and vehicle-side sensors, complicating data synchronization and reducing reliability. CRUISE, the first simulation framework capable of reconstructing real-world street scenes and generating synchronized V2X data from both ego-vehicle and infrastructure views—without any temporal offset. This precise alignment improves data accuracy and paves the way for more reliable real-world V2X deployment.
\subsection{Accurate 3D Boxes.}
Annotation errors in V2X-Seq's 3D bounding boxes can introduce bias into models during real-world deployment. CRUISE addresses this issue by leveraging Street Gaussians for improved foreground–background decoupling, thereby optimizing 3D box annotations. It further ensures annotation accuracy through controlled scene editing during data generation. As a result, CRUISE not only produces high-fidelity synthetic data but also significantly reduces labeling errors—enhancing model reliability and improving real-world performance.

\section{Limitations}
While CRUISE could generate highly accurate and realistic V2X data, it still has several limitations. When the viewpoints of the ego-vehicle and infrastructure are directionally similar in the Bird’s Eye View (BEV) perspective, LiDAR noise can introduce artifacts, causing road textures to appear mid-air in the ego-car view during rendering. Future work may explore filtering techniques to mitigate this issue.Although CRUISE is capable of generating point clouds, it does not simulate realistic LiDAR point clouds for both ego-vehicle and infrastructure perspectives. Future work may incorporate the visibility mask approach from Occ3D \cite{tian2023occ3d} for this purpose. Furthermore, the GS method used in this work struggles with rainy scenes, particularly when water droplets obscure the ego-camera view. Future work could explore training a diffusion model, inspired by DeRainGS \cite{liu2024deraings}, to enhance image quality in adverse weather conditions.

\section{Conclusion}
In this work, we present CRUISE, the first V2X driving simulator based on dynamic GS, equipped with powerful editing and realistic rendering capabilities to generate V2X data. From the V2X data collected in the real world, we can reconstruct the environment and flexibly place various vehicles in the scene under the guidance of a traffic flow generator. With the rendering technique, we could generate new V2X datasets for downstream tasks. Extensive experiments demonstrate that CRUISE could improve tasks like 3D detection across ego-vehicle, infrastructure, and cooperative views, as well as cooperative 3D tracking. In addition, CRUISE enables the generation of diverse corner cases and high-quality samples, contributing to a closed-loop data ecosystem for training and evaluation in future V2X autonomous driving systems. These capabilities support the development of more robust and scalable end-to-end autonomous driving technologies.

\bibliographystyle{IEEEtran}
\balance
\bibliography{ref}
\end{document}